\documentclass{article}

\usepackage{arxiv}

\usepackage[utf8]{inputenc} 
\usepackage[T1]{fontenc}    
\usepackage{hyperref}       
\usepackage{url}            
\usepackage{booktabs}       
\usepackage{amsfonts}       
\usepackage{nicefrac}       
\usepackage{microtype}      
\usepackage{lipsum}
\usepackage{graphicx}
\graphicspath{ {./images/} }
\usepackage{amsmath}

\title{TYolov5: A Temporal Yolov5 Detector Based on Quasi-Recurrent Neural Networks for Real-Time Handgun Detection in Video}

\author{
  Mario Alberto Duran-Vega \\
  School of Engineering and Science\\
  Tecnologico de Monterrey\\
  Monterrey, NL 64849, MEXICO \\
  \texttt{A00755076@exatec.tec.mx} \\
   \And
 Miguel Gonzalez-Mendoza \\
  School of Engineering and Science\\
  Tecnologico de Monterrey\\
  Lopez Mateos, Mex 52926, MEXICO\\
  \And
  Leonardo Chang \\
  School of Engineering and Science\\
  Tecnologico de Monterrey\\
  Lopez Mateos, Mex 52926, MEXICO \\
    \And
 Cuauhtemoc Daniel Suarez-Ramirez \\
  School of Engineering and Science\\
  Tecnologico de Monterrey\\
  Monterrey, NL 64849, MEXICO \\
}

\begin{document}
\maketitle
\begin{abstract}
Timely handgun detection is a crucial problem to improve public safety; nevertheless, the effectiveness of many surveillance systems still depends of finite human attention. Much of the previous research on handgun detection is based on static image detectors, leaving aside valuable temporal information that could be used to improve object detection in videos. To improve the performance of surveillance systems, a real-time temporal handgun detection system should be built. Using Temporal Yolov5, an architecture based on Quasi-Recurrent Neural Networks, temporal information is extracted from video to improve the results of handgun detection. Moreover, two publicly available datasets are proposed, labeled with hands, guns, and phones. One containing 2199 static images to train static detectors, and another with 5960 frames of videos to train temporal modules. Additionally, we explore two temporal data augmentation techniques based on Mosaic and Mixup. The resulting systems are three temporal architectures: one focused in reducing inference with a mAP$_{50:95}$ of 55.9, another in having a good balance between inference and accuracy with a mAP$_{50:95}$ of 59, and a last one specialized in accuracy with a mAP$_{50:95}$ of 60.2. Temporal Yolov5 achieves real-time detection in the small and medium architectures. Moreover, it takes advantage of temporal features contained in videos to perform better than Yolov5 in our temporal dataset, making TYolov5 suitable for real-world applications. The source code is publicly available
at https://github.com/MarioDuran/TYolov5.
\end{abstract}

\keywords{Computer vision \and ConvLSTM \and QRNN \and Real-time \and Temporal Object Detection \and Yolov5}

\section{Introduction}
In recent years, Deep Learning (DL) has become very popular in the field of Computer Vision (CV) \cite{Lecun2015deep}. Convolutional Neural Networks (CNNs) are one of the main DL techniques for CV \cite{olmos2018automatic}; with the aid of the ever-increasing computational power and the GPUs, they have improved the performance of CV domain-specific tasks like object detection \cite{jiao2019survey}.

Many surveillance cameras remain a passive video tool for detecting crime at stores or streets. To increase their effectiveness, the development of automated object detection tools, such as handgun detectors, might help to fulfill their purpose. 

Object detection falls mainly into two categories: one stage, and two stage detectors  \cite{jiao2019survey}. Nevertheless, although one stage detectors achieve real-time detection in video, they are designed to work with static images; therefore, they do not take advantage of temporal information contained in videos, which could be used to improve their results.

Recurrent neural networks (RNNs), have been extensively used when the input data is sequential and contain valuable temporal information, such as: text, stock data, audio, and video \cite{yu2019review}. However, RNNs are unable to learn if the gap of the input is too large  \cite{yu2019review}. To better handle the problem of long-term dependencies, Long Short-Term Memory (LSTM) \cite{hochreiter1997long} was proposed. Since then, almost every interesting result based in RNNs, have been accomplished with the aid of LSTMs \cite{yu2019review}. Some of those examples are temporal object detectors that are built upon still-image detectors, such as TSSD-OTA \cite{chen2019temporally} and Mobile video object detection \cite{liu2018mobile}; approaches that take advantage of spatio-temporal features, using RNNs implemented as LSTMs.

LSTMs are implemented to extract temporal features of sequential information, but for the case of videos, the sequential information comes with spatial information as well. In response to this problem, Convolutional LSTM (ConvLSTM) was proposed to extract spatio-temporal features from sequential images \cite{xingjian2015convolutional}. 

However, the accuracy obtained by using ConvLSTMs is directly related with the quality of the spatio-temporal features they receive \cite{chen2019temporally}. As result, approaches like attention modules have become popular in RNNs \cite{Liu_2017_CVPR, chen2019temporally, wang2016attention}, which allow models to focus on certain information and improve their results. 

RNNs are a powerful tool for extracting information from sequential data. Nonetheless, the dependence of the previous step to calculate the current one, makes training slow and prevents it from taking advantage of the parallelism offered by GPUs. Quasi-recurrent neural networks (QRNNs) \cite{gehring2017convolutional, wei20203} use convolutional layers to apply parallel feature extraction across time steps. They use a recurrent pooling function to produce the final output. With this type of architecture, QRNNs claim to speed up training by 16 times, whereas obtaining better results than RNNs for specific tasks. Only being less accurate when the task requires extracting long-distance context features.

To the best of our knowledge QRNNs have never been applied to temporal object detection. This is the first work that extracts spatio-temporal features with QRNNs to address the problem of handgun detection. Results show that QRNNs achieve comparable results to ConvLSTMs in terms of accuracy, but with faster inference time when using full precision floats.

\subsection{Contributions and Proposed Approach}

\begin{figure}[!ht] 
    \centering
    \includegraphics[width=0.4\textwidth]{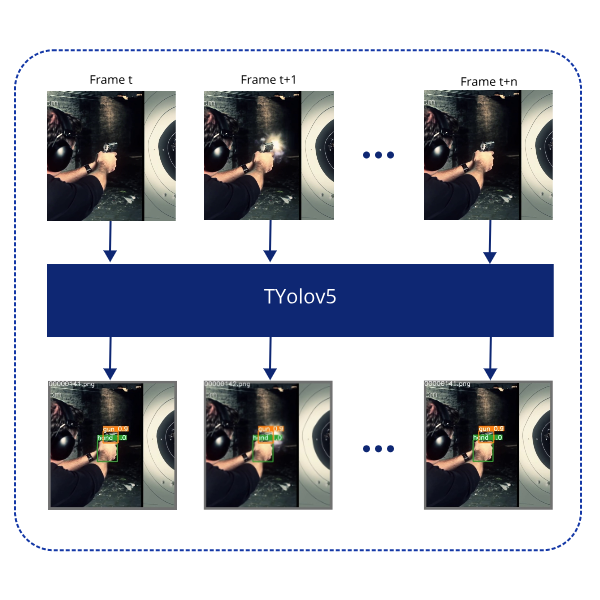}
    \caption{Yolov5 improves object detection in videos by taking advantage of spatio-temporal features and maintaining frames per second above 30.}
    \label{fig:TYolov5_demo}
\end{figure}

As shown in the Fig. \ref{fig:TYolov5_demo}, our model can detect objects in video using QRNNs. Our temporal detector system is based on Yolov5, namely, TYolov5, and integrates multi-scale temporal features using QRNN modules.

Object detection at different scales remains a challenging task in CV. Thus, pyramidal architectures were proposed; some successful examples implementing this approach are the most recent Yolo family \cite{redmon2018yolov3, bochkovskiy2020yolov4, jocher}. The purpose of pyramidal architectures is to produce multi-scale feature maps to improve the detection of small and large objects at their scale. Yolov5 follows this logic, and generates feature maps at three scales; therefore, we built a QRNN module to take advantage of the information obtained at each scale, and their temporal propagation. All the aforementioned characteristics concede TYolov5 an improvement in accuracy over Yolov5 in our temporal dataset, whereas keeping the required performance of a real-time detector. Our contributions are summarized as follows:

\begin{itemize}
    \item Two datasets are publicly available. One containing 2199 labeled images of hands, guns, and phones (1989 for training and 210 for testing). And another containing 5960 labeled images of hands, guns, and phones as well (50 videos of 100 frames for training, and 48 videos of 20 frames for testing).
    
    \item A temporal module based in QRNNs with the purpose of propagate multi-scale feature maps through time. This module lets us take advantage of the parallelism offered by GPUs, and reduces inference time. 
    
    \item TYolov5, a real-time temporal object detector suitable for real-world applications.
    
    \item Based in the mosaic augmentation created by \cite{jocher}, and mixup augmentation from \cite{zhang2017mixup}. We propose, temporal mosaic augmentation and temporal mixup respectively. 
    
\end{itemize}

This paper is organized as follows: section \ref{sc:related} provides a detailed summary of related works; section \ref{sc:approach} describes the presented architecture, its sub-modules and temporal data augmentation techniques; section \ref{sc:experiments} shows the datasets, the experimental results and discussion; finally section \ref{sc:conclusion}, presents the conclusion and the future work.

\section{Related works}
\label{sc:related} 

\subsection{Object Detection}

A fundamental and challenging problem in computer vision is object detection \cite{liu2020deep}. Given some categories (like phones, hands or computers), the objective is to determine if such categories exist within the image, and if they do, return the spatial location of every object detected \cite{liu2020deep}. Recently, the most notable contributions regarding object detection, are associated with the development of deep convolutional neural networks and the power of GPUs \cite{jiao2019survey}. Deep learning has been adopted to the point where almost every state-of-the-art object detector uses deep learning networks as feature extractor (backbone) and detector (head) \cite{jiao2019survey}.

\subsubsection{One-stage vs Two-stage Detectors}

Deep learning approaches for object detection, fall mainly into two categories: one-stage detectors, and two-stage detectors \cite{jiao2019survey}. One-stage detectors achieve high inference speeds, in contrast to two-stage detectors, which have high accuracy at the cost of speed. The most representative of the second category is the R-CNN Family \cite{girshick2014rich}: fast R-CNN \cite{girshick2015fast}, faster R-CNN \cite{ren2015faster}, R-FCN \cite{dai2016r}, and Libra R-CNN \cite{pang2019libra}. The R-CNN Family, first uses a region proposal network to estimate regions of interest, then, based on the proposed regions, performs object classification and the estimation of the bounding boxes, hence the two-stage category. On the other hand, one-stage detectors, which most popular architectures are: RetinaNet \cite{lin2017focal}, SSD \cite{liu2016ssd} and the YOLO Family \cite{redmon2016you,redmon2017yolo9000,redmon2018yolov3, bochkovskiy2020yolov4, jocher}, are models which approach object detection as a regression problem. Simultaneously calculating the class probabilities and the bounding boxes, thus, the one-stage category.
One benefit of building faster one-stage detectors like Yolo, is that they can be used as real-time object detectors in video \cite{zhang2020video}, allowing computers to achieve tasks such as driving cars or surveilling an area of interest \cite{redmon2016you}.

\subsubsection{Video Object Detection}

Still-image object detectors can be implemented as video object detectors. However, rich temporal information would be wasted, since still-image based detectors would treat each sequence frame individually \cite{zhang2020video}. With the purpose of designing object detectors that could take advantage of temporal information, video object detection challenge was introduced by ImageNet Large Scale Visual Recognition (LSVRC2015), attracting considerable research attention since then \cite{zhang2019modeling, russakovsky2015imagenet}.

The first winning method was T-CNN \cite{kang2017t}, taking advantage of temporal and contextual information using tubelets, it improved the base performance of static images detectors. Another approach was MANet \cite{wang2018fully} which uses pixel and instance-level calibration across frames, to obtain better results on video object detection. Additional temporal approaches were: D\&T \cite{feichtenhofer2017detect}, an architecture for simultaneous tracking and detection; STMN \cite{xiao2018video}, which introduces spatial-temporal memory networks to model long-term temporal appearance and dynamics; and Seq-NMS \cite{han2016seq}, a post-processing proposal which uses high-scoring objects detection from adjacent frames to boost the scores of weaker detections. 

All the aforementioned approaches are non-causal, and to produce an output for frame $t$, they require future frames $t+n$. This condition makes them unsuitable to work with a continuous stream of frames provided by real-time surveillance systems, since $t+n$ frames do not exist at the moment a frame $t$ is displayed in video. 


\subsubsection{LSTM-based causal video object detection}

Causal video detectors have received less research attention \cite{zhang2020video}, some few approaches are: TSSD \cite{chen2019temporally} a temporal causal detector which extract feature maps at different scales with SSD, then uses attentional ConvLSTMs to propagate temporal information and improve the object detection; Temporal MobileNet \cite{liu2018mobile}, proposes an online model based in MobileNet \cite{howard2017mobilenets}, which uses Bottleneck-LSTMs to refine and propagate feature maps across time; Flow\&LSTM \cite{zhang2019modeling}, uses ConvLSTMs with sequences of only two frames to improve the object detection, but instead of using an attention module, they use FlowNet \cite{dosovitskiy2015flownet} to select the most important features. 

However, the aforementioned causal video object detectors are based on LSTMs, which makes inference slow when compared to QRNNs. Another disadvantage is that they only propose a single architecture. But depending on the application, the user might have different requirements, such as: prioritize accuracy over FPS, FPS over accuracy, or a balance between them.

\subsubsection{Gun Detection}

Surveillance systems still require human intervention to detect ongoing armed robberies or other crimes. One of the first deep learning approaches related to automatic gun detection in video, is probably the work of Olmos et al. \cite{olmos2018automatic}. It approaches handgun detection as a static image detection problem, and it is solved by using a two-stage detector (Faster R-CNN). Other handgun detection approaches implemented their solution by using two-stage detectors \cite{olmos2019binocular, raturi2019adocw, fernandez2019gun, gonzalez2020real}. Recent solutions used one-stage detectors from the Yolo Family \cite{abruzzo2019cascaded, romero2019convolutional, warsi2019gun}, and alternative architectures like M2Det \cite{lim2019gun}. Nevertheless, all the aforementioned approaches neglect to take advantage of temporal information available in videos, which could be used to improve the gun detection (Table \ref{table:Handgun_works}). 

\begin{table}[!ht]
\centering
\caption{Principal works about handgun detection. Listed approaches do not use temporal information.}
\begin{tabular}{@{}ll@{}}
\toprule
Gun Detection Work                          & Approach     \\ \midrule
\cite{olmos2018automatic} Olmos et al. 2018    & Faster R-CNN     \\
\cite{olmos2019binocular} Olmos et al. 2019    & Faster R-CNN     \\
\cite{raturi2019adocw} Raturi et al. 2019        & Faster R-CNN     \\
\cite{fernandez2019gun} Fernandez-Carrobles et al 2019       & Faster R-CNN     \\
\cite{gonzalez2020real} Gonzalez et al. 2020       & Faster R-CNN-FPN \\
\cite{abruzzo2019cascaded} Abruzzo et al. 2019    & Yolov3           \\
\cite{romero2019convolutional} Romero et al. 2019 & Yolov3           \\
\cite{warsi2019gun} Warsi et al. 2019          & Yolov3           \\
\cite{lim2019gun} Lim et al. 2019              & M2Det            \\ \bottomrule
\end{tabular}
\label{table:Handgun_works}
\end{table}

\section{Approach}
\label{sc:approach} 

Yolov5 claims to have comparable results to Yolov4 in COCO dataset \cite{bochkovskiy2020yolov4}, whereas reducing the training time by 52 times for custom datasets \cite{jocher}.
Since training temporal object detectors is slower than training static object detectors, and Yolov5 and Yolov4 achieve better results in COCO dataset, than all the architectures used by the state-of-the-art for handgun detection (Table \ref{table:Handgun_works}), we chose Yolov5 as base for our temporal architecture. 

\subsection{TYolov5}

\begin{figure}[!ht]
    \centering
    \includegraphics[width=1\textwidth]{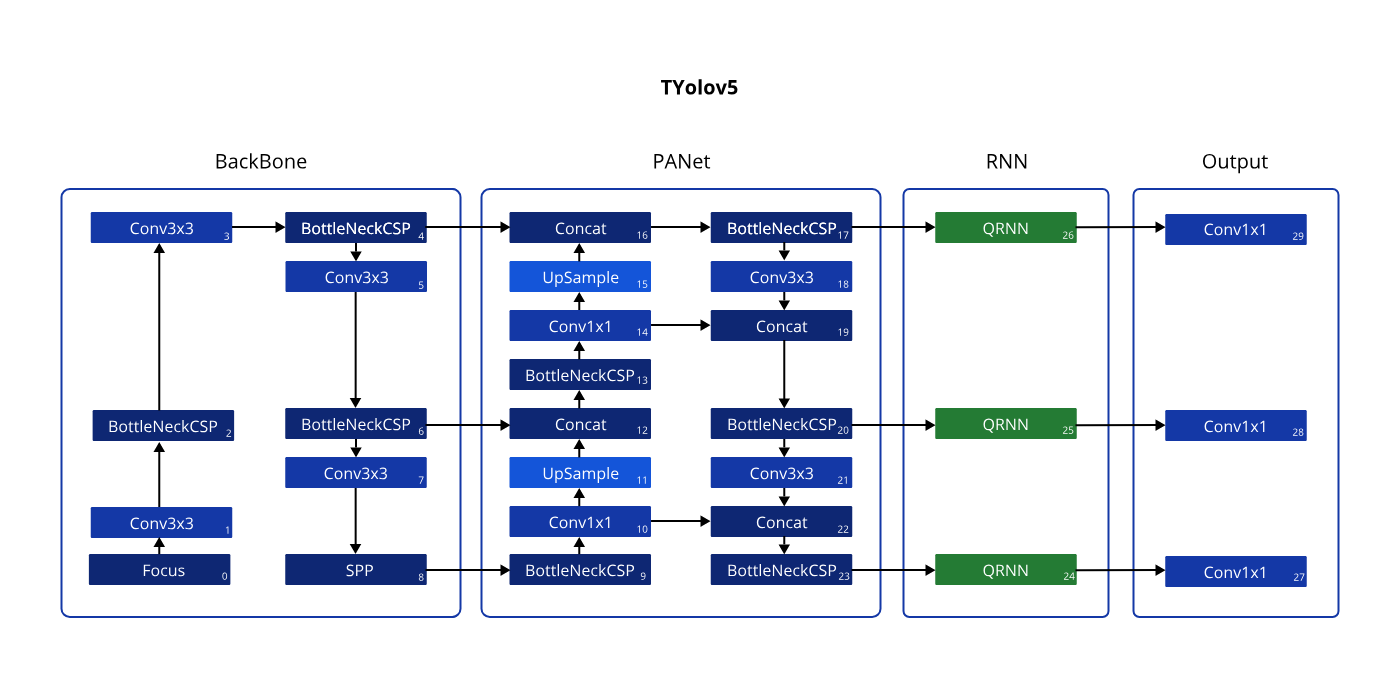}
    \caption{Architecture of TYolov5. The backbone is composed by BottleNeckCSPs, which perform feature extraction at three scales (Low, Mid, High), the selected features are passed to PANet for feature aggregation. Next, each of the three outputs is passed to a quasi-recurrent neural network (QRNN) to extract spatio-temporal features. Finally, each output of the QRNN is passed to the detection head to perform object detection.}
    \label{fig:TYolov5}
\end{figure}

The proposed architecture is build upon Yolov5 \cite{jocher} ( Fig. \ref{fig:TYolov5}), to extend it to support temporal information, we append a RNN module to each of the three outputs of PANet neck. Our proposed architecture is built in four blocks.

\begin{figure}[!ht] 
    \centering
    \includegraphics[width=0.4\textwidth]{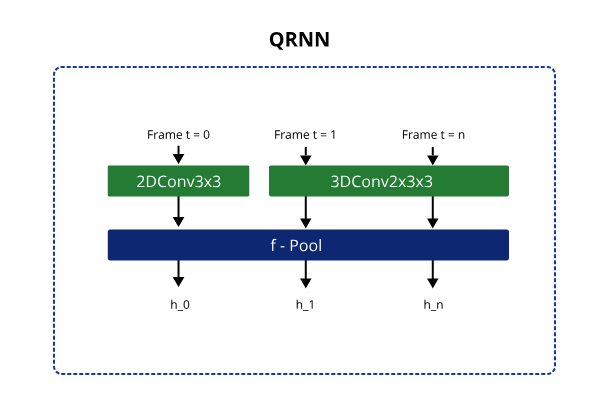}
    \caption{QRNN module implemented with a 2D Convolution to generate $h_0$, then a 3D Convolution to extract spatio-temporal features of two or more frames. Finally, the F-pool layer to mix the hidden states.}
    \label{fig:qrnn_module}
\end{figure}

\begin{itemize}
    \item \textbf{Backbone:} Based on BottleNeckCSPs to generate the pyramidal scaled features for object detection.
    \item \textbf{Neck:} Built upon PANet, which is the module in charge of feature aggregation.
    \item \textbf{RNN:} Appending a QRNN module (Fig. \ref{fig:qrnn_module}), at each of the three outputs of PANet.
    \item \textbf{Head:} Yolov3 Head, like Yolov4 and Yolov5. 
\end{itemize}

Like Yolov3, the final convolution of Yolov5 is responsible for object prediction with the multi-scaled features received from the pyramidal architecture. Next, it generates a fixed number of bounding boxes and category scores to determine if an object exists within the processed frame. However, sometimes redundant bounding boxes are predicted by the network. Thus, Non-Maximum Suppression (NMS) algorithm is applied to determine the most relevant bounding boxes from the image. 

We define three TYolov5 architectures to perform our experiments: TYolov5 small, TYolov5 medium, and TYolov5 large. TYolov5 small, aims to reach the highest FPS among all TYolov5 architectures. TYolov medium, aims to have a balance between FPS and the mAP. Finally, TYolov5 large, aims to achieve the highest mAP in our temporal dataset. Each of the models are implemented with QRNNs and ConvLSTMs to compare their effectiveness within the temporal handgun detection domain. 

\subsubsection{QRNN TYolov5 Small}

For this network, we define: an image resolution of $640 \times 640$, a batch size of 40, and a sequence size of 2 video frames. Next, the backbone of this architecture: BottleneckCSP$_4$, BottleneckCSP$_6$, and SPP$_8$. Produces pyramidal features of size [batch size $\times$ sequence size, 128, 80, 80], [batch size $\times$ sequence size, 256, 40, 40], [batch size $\times$ sequence size, 512, 20, 20] respectively.
Following, PANet take the aforementioned pyramidal outputs to aggregate features, outputting features of size [batch size $\times$ sequence size, 128, 80, 80], [batch size $\times$ sequence size, 256, 40, 40], [batch size $\times$ sequence size, 512, 20, 20] from the modules BottleneckCSP$_{17}$, BottleneckCSP$_{20}$ and BottleneckCSP$_{23}$ respectively. Then, TYolov5 uses its three temporal modules, namely QRNNs, to propagate temporal information through the model. Taking as input the aforementioned features from the neck, and outputting feature tensors of dimensions [batch size $\times$ sequence size, 128, 80, 80], [batch size $\times$ sequence size, 256, 40, 40], [batch size $\times$ sequence size, 512, 20, 20] from the modules QRNN$_{24}$, QRNN$_{25}$ and QRNN$_{26}$. The QRNN module illustrated in Fig. \ref{fig:qrnn_module}, performs a 2D convolution to create the first hidden state, and a 3D convolution to calculate the following hidden states. The procedure executed within the QRNN is illustrated in \eqref{eq:qrnn}.

\begin{align}
\begin{split}\label{eq:qrnn}
    z_{t}={}&tanh(W^{1}_{x}x_{t-1} + W^{2}_{z}x_{t}). \\
    f_{t}={}&sigmoid(W^{1}_{f}x_{t-1} + W^{2}_{f}x_{t}). \\
\end{split}
\end{align}

Inspired in the element-wise gated architectures found in LSTMs, F-Pool \eqref{eq:qrnn_fo} mixes the hidden states across time steps.

\begin{align}
\begin{split}\label{eq:qrnn_fo}
    h_t={}&f_t \odot c_{t-1} + i_t \odot z_t. \\
\end{split}
\end{align}

We pass the resulting tensors to Yolov5 detector head, which is the anchor based head implemented in yolov4, and yolov3. Finally, we perform the final prediction of the bounding boxes.

\subsubsection{ConvLSTM TYolov5 Small}

The ConvLSTM is an extended version of the LSTM, which was first presented by Xingjian et al. \cite{xingjian2015convolutional} with the purpose of learning spatio-temporal features for the problem of precipitation nowcasting. For TYolov5 small, we replace the QRNN module with the ConvLSTM model implemented in TSSD \cite{chen2019temporally}, and since we are aiming to obtain the best inference time possible, the attention module for this ConvLSTM is removed \eqref{eq:minimal_convlstm}.

\begin{align}
\begin{split}\label{eq:minimal_convlstm}
    i_{t}={}&sigmoid(W_{i}*[x ,a_{t} \circ h_{t-1}] + b_{i}). \\
    f_{t}={}&sigmoid(W_{f}*[x ,a_{t} \circ h_{t-1}] + b_{f}). \\
    o_{t}={}&sigmoid(W_{o}*[x, a_{t} \circ h_{t-1}] + b_{o}). \\
    c_{t}={}&sigmoid(W_{c}*[x,a_{t} \circ  h_{t-1}] + b_{c}). \\
    s_{t}={}&(f_{t} \odot s_{t-1}) + (i_t \odot c_t). \\
    h_{t}={}&O_{t} \odot tanh(S_{t}). \\
\end{split}
\end{align}

The definition of symbols is as follows: $*$ represents a convolution, $[x,y]$ denotes concatenation of x and y, $ \circ$ represents channel wise multiplication, $\odot$ constitutes an element-wise multiplication, $t$ represents time, $i_{t}$ the input gate, $f_{t}$ forget gate, $o_{t}$ output gate, the incoming information $c_{t}$, memory $s_{t}$, and $h_{t}$ the hidden state.

\subsubsection{QRNN TYolov5 Medium}

To produce a more robust Yolov5 architecture that achieves higher accuracy, we need to increase the number of channels supported by each module. To build a medium size TYolov5, we define: an image resolution of $640 \times 640$, a batch size of 36, and a sequence size of 2 video frames. Next, the backbone of this architecture, BottleneckCSP$_4$, BottleneckCSP$_6$, and SPP$_8$, is going to produce pyramidal features of size [batch size $\times$ sequence size, 192, 80, 80], [batch size $\times$ sequence size, 384, 40, 40], [batch size $\times$ sequence size, 768, 20, 20] respectively.
Following, PANet take the aforementioned pyramidal outputs to aggregate features, outputting features of size [batch size $\times$ sequence size, 192, 80, 80], [batch size $\times$ sequence size, 384, 40, 40], [batch size $\times$ sequence size, 768, 20, 20] from the modules BottleneckCSP$_{17}$, BottleneckCSP$_{20}$ and BottleneckCSP$_{23}$ respectively. Then, TYolov5 uses three QRNNs to propagate temporal information. Taking as input the aforementioned features from the neck, and outputting feature tensors of dimension  [batch size $\times$ sequence size, 192, 80, 80], [batch size $\times$ sequence size, 384, 40, 40], [batch size $\times$ sequence size, 768, 20, 20] from the modules QRNN$_{24}$, QRNN$_{25}$ and QRNN$_{26}$. Then, the final prediction of bounding boxes is processed by the Yolov5 head.

\subsubsection{ConvLSTM TYolov5 Medium}

To build a TYolov5 medium ConvLSTM architecture, we repeat the same procedure as before. We replace the QRNN module from TYolov5 medium, with a ConvLSTM. This is one of the key advantages of our architecture, each module can be interchanged with another designed for the same purpose. 

\subsubsection{QRNN TYolov5 large}

We increase the number of filters to build a more robust temporal object detector. We define: an image resolution of $640 \times 640$, a batch size of 28, and a sequence size of 2 video frames. Next, the backbone of this architecture: BottleneckCSP$_4$, BottleneckCSP$_6$, and SPP$_8$. Is going to produce pyramidal features of size [batch size $\times$ sequence size, 256, 80, 80], [batch size $\times$ sequence size, 512, 40, 40], [batch size $\times$ sequence size, 1024, 20, 20] respectively.
Following, PANet take the aforementioned pyramidal outputs to aggregate features, outputting features of size [batch size $\times$ sequence size, 256, 80, 80], [batch size $\times$ sequence size, 512, 40, 40], [batch size $\times$ sequence size, 1024, 20, 20] from the modules BottleneckCSP$_{17}$, BottleneckCSP$_{20}$ and BottleneckCSP$_{23}$ respectively. Then, TYolov5 uses three QRNNs to propagate temporal information. Taking as input the aforementioned features from the neck, and outputting feature tensors of dimension  [batch size $\times$ sequence size, 256, 80, 80], [batch size $\times$ sequence size, 512, 40, 40], [batch size $\times$ sequence size, 1024, 20, 20] from the modules QRNN$_{24}$, QRNN$_{25}$ and QRNN$_{26}$. Then, the final prediction of bounding boxes is processed by the Yolov5 head.

\subsubsection{ConvLSTM TYolov5 Large}

Following the same logic, we change the QRNN module of the large model of TYolov5, for a ConvLSTM which is capable of handling the same number of channels as the replaced QRNN.


\subsection{Temporal Data Augmentation}

To train TYolov5, we kept the static data augmentation used to train Yolov5. The techniques used were: hsv color modification, scale modification, translate, shear modification, perspective modification, and rotation. Additionally, to train TYolov5 with our temporal dataset, we implemented five temporal data augmentation techniques.

\subsubsection{Temporal Mosaic}
One of the data augmentation techniques implemented in Yolov4 and Yolov5, was mosaic data augmentation. This approach combines four training contexts into one single image. Moreover, batch normalization is performed with different contexts on each layer, reducing the need for large batch sizes. Mosaic augmentation improved the $AP_{50}$ of Yolov4 by 1.8\% in COCO Dataset, proving that this data augmentation technique is worth exploring for temporal object detectors \cite{bochkovskiy2020yolov4}.

\begin{figure}[!ht] 
    \centering
    \includegraphics[width=0.4\textwidth]{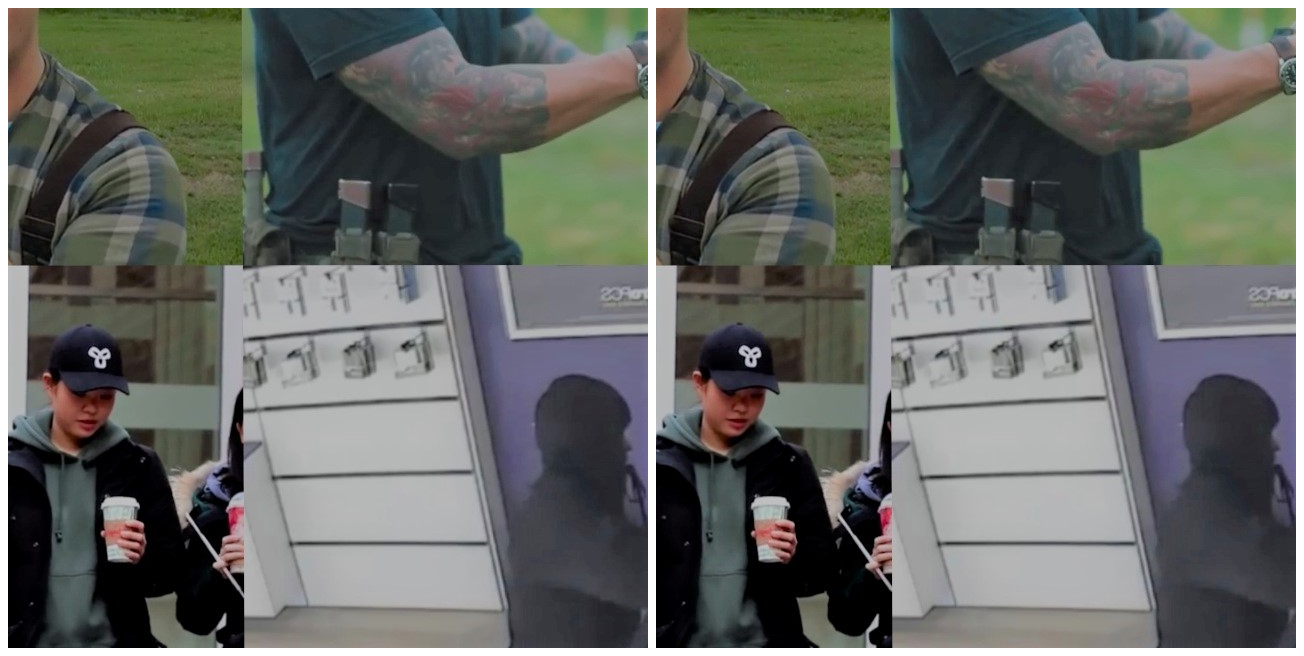}
    \caption{Temporal Mosaic Augmentation. It concatenates four videos with frames at time $t$, then for the next image in the sequence, concatenates frames at time $t+1$.}
    \label{fig:tmosaic}
\end{figure}

Training spatio-temporal RNNs with a small dataset is a difficult task since they quickly tend to overfit. To that end, we propose a temporal data augmentation technique inspired in mosaic augmentation \cite{jocher}. Instead of images, Temporal Mosaic concatenates four videos in which frames are played along with the sequence time (Fig. \ref{fig:tmosaic}).

\subsubsection{Temporal Mixup}
Mixup is data augmentation technique which has shown a generalization improvement for some architectures in datasets like ImageNet-2012, CIFAR-10 and CIFAR-100 \cite{zhang2017mixup}. Mixup consists in mixing two different images and their bounding boxes into one, combining two contexts to regularize the neural network.

\begin{figure}[!ht] 
    \centering
    \includegraphics[width=0.4\textwidth]{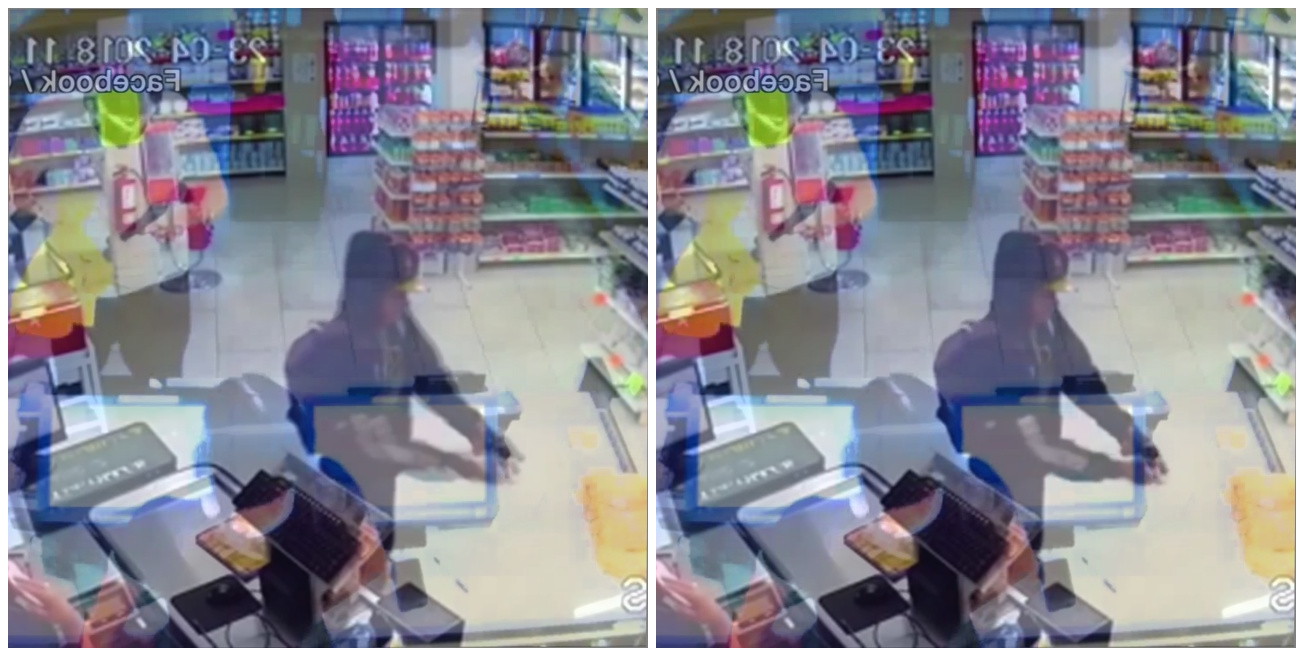}
    \caption{Temporal Mixup Augmentation. It mixes two frames at time $t$ from two video sequences. Then, for the next frame, it mixes two frames from time $t+1$, from the same video sequences.}
    \label{fig:tmixup}
\end{figure}

To improve the generalization of our temporal architecture, we propose temporal mixup. however, instead of mixing two images, we mix two frames of a video at time $t$ with their corresponding bounding boxes. (Fig. \ref{fig:tmixup}).

\subsubsection{Random Erasing} 

\begin{figure}[!ht]
\centering
  \includegraphics[width=0.4\textwidth]{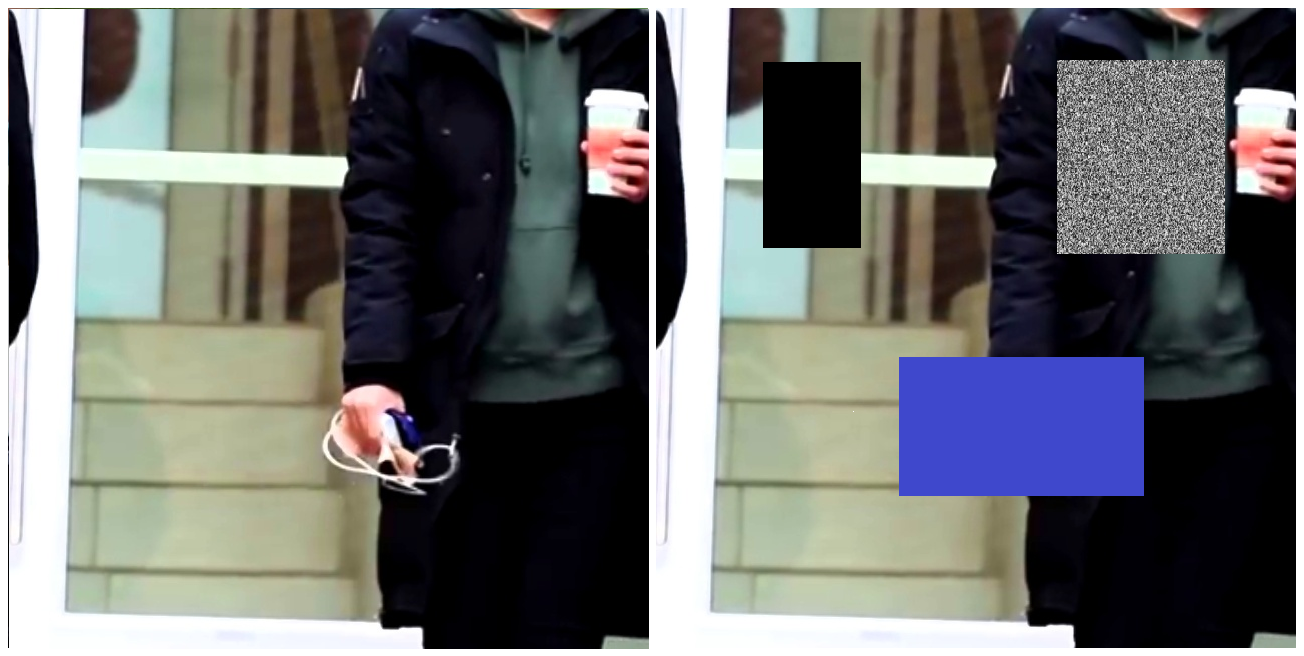}
  \caption{Random Erasing. It adds a random rectangle to the picture with a uniform random color, or salt and pepper noise.}
  \label{fig:randomerasing}
\end{figure}

One common situation in video object detection is to detect objects that are occluded by another object. Random Erasing augmentation \cite{zhong2020random} can help to simulate those situations and occlude an object which previously was detected. We added random erasing to each frame of the video sequence with the exception of the first frame. This for preventing the network of learning patterns from occluded objects (Fig. \ref{fig:randomerasing}). 

\subsubsection{Random Blur} 
\begin{figure}[!ht]
\centering
  \includegraphics[width=0.4\textwidth]{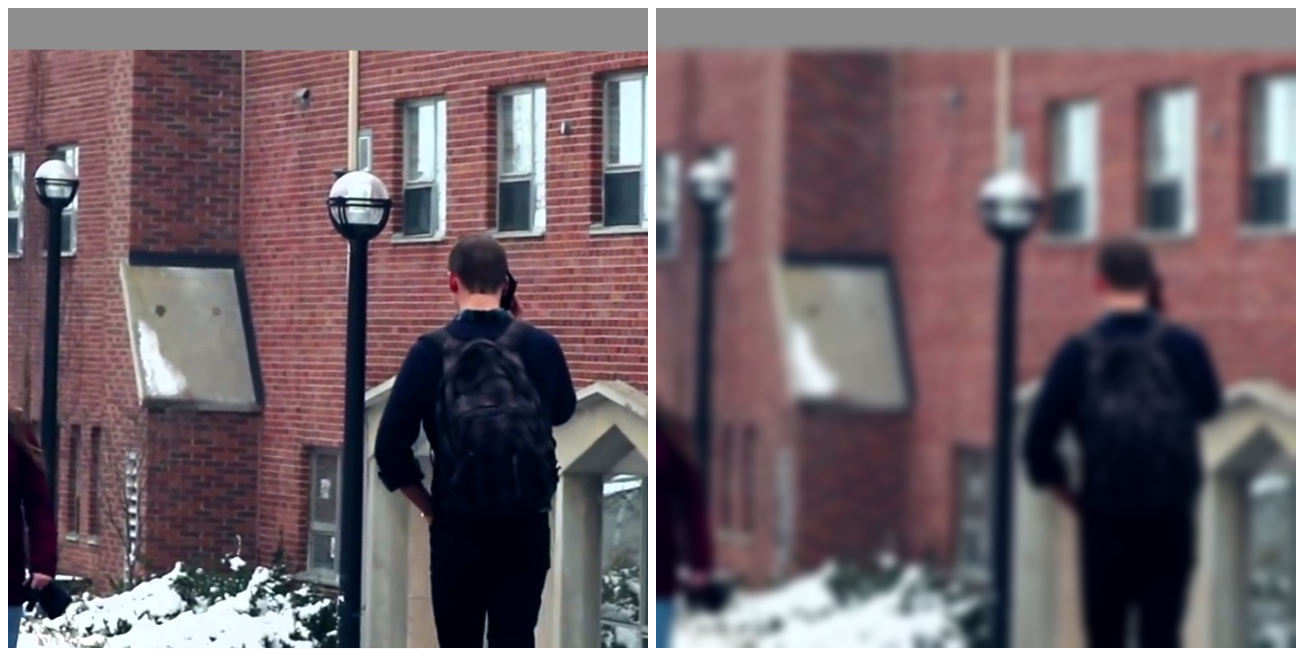}
  \caption{Random Blur. It randomly adds Gaussian blur to the image to simulate focal changes in the camera. }
  \label{fig:randomblur}
\end{figure}

When working with video data streams, the camera usually suffers from random focal changes. This adds blur to images, and prevents object detectors from extracting meaningful features to perform an accurate detection. To still be able to recognize objects under that blur, the detector should use information from previous frames. To prepare our temporal object detector for these kinds of scenarios, we simulate them by adding a random blur to each frame when training our temporal object detector (Fig. \ref{fig:randomblur}).

\subsubsection{Gaussian noise} 
\begin{figure}[!ht]
\centering
  \includegraphics[width=0.4\textwidth]{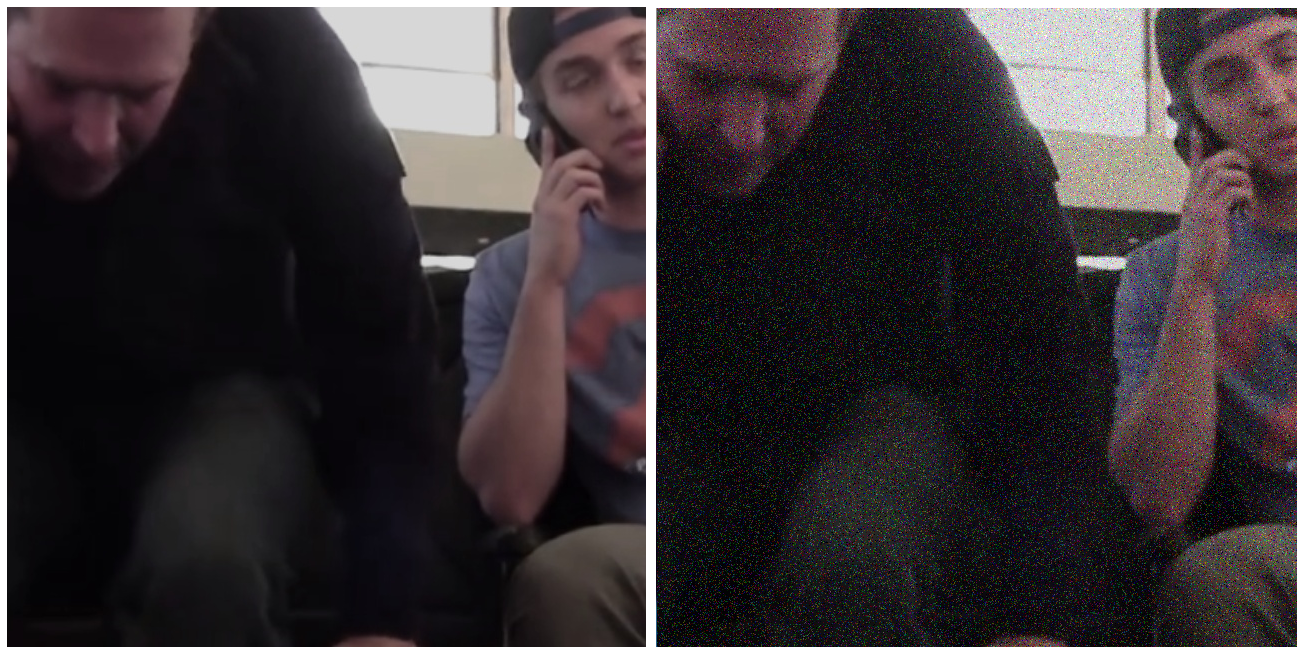}
  \caption{Random Gaussian Noise. It simulates common sensor noise in videos. }
  \label{fig:randomgn}
\end{figure}

Gaussian noise is a common problem when working with video. Noise can reduce the accuracy of models, and lead to less generalization power when working with real-world data. To prepare our temporal object detector to work under this context, we randomly added Gaussian noise to each frame to train our temporal object detector for these kinds of scenarios (Fig. \ref{fig:randomgn}).

\section{Experiments and Discussion}
\label{sc:experiments} 

\subsection{Datasets}
\subsubsection{Hands Guns and Phones (HGP) Dataset}
\begin{figure}[!ht] 
    \centering
    \includegraphics[width=0.5\textwidth]{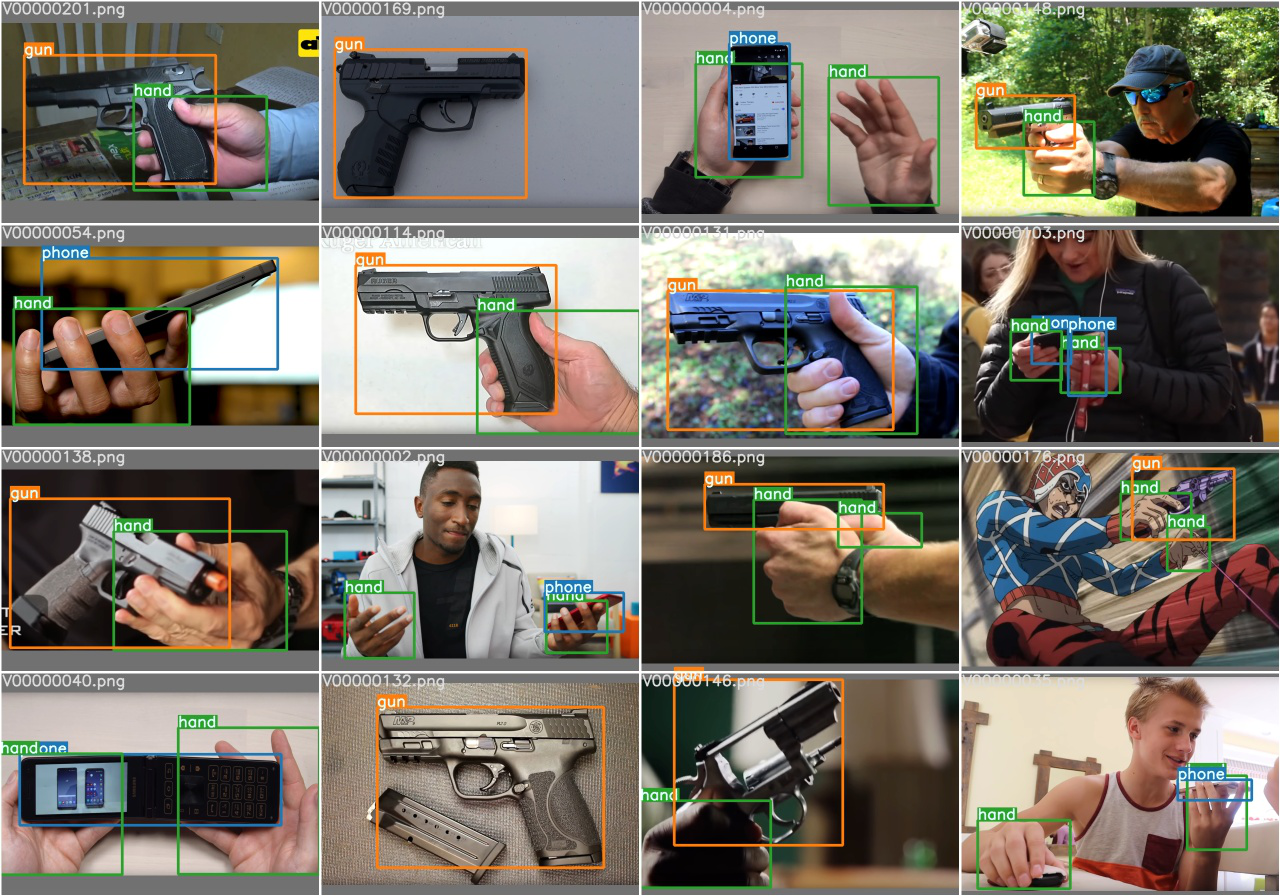}
    \caption{Demo images of our THGP and HGP dataset.}
    \label{fig:datasets}
\end{figure}

 According to Olmos et al. \cite{olmos2018automatic} the current datasets for handgun detection contain two possible enhancements: handgun datasets do not contain contain multiple categories to let the network learn the differences between similar features; The dataset contain images of handguns which are not used in context. Following those recommendations, we compiled the Hands Guns and Phones (HGP) dataset. It contains 2199 images (1989 for training an 210 for testing) of people using guns or phones in real-world scenarios (people making phones reviews, shooting drills, or making calls). Every image of this dataset is labeled with the bounding boxes of Hands, Phones and Guns. All the aforementioned images were collected from Youtube videos and have different sizes (Fig. \ref{fig:datasets}). This dataset is available in google drive: \href{https://drive.google.com/file/d/138Zp7MuchcS4He6LBFSTow5q97BwnpWv/view?usp=sharing}{\textbf{HGP}} or go to the next url: \url{https://drive.google.com/file/d/138Zp7MuchcS4He6LBFSTow5q97BwnpWv}.

\subsubsection{Temporal Hands Guns and Phones (THGP) Dataset}
Temporal Hands Guns and Phones (THGP) dataset, is a collection of 5960 video frames (5000 for training and 960 for testing). The training part is composed with 50 videos of 100 frames ($720\times720$ pixels). This dataset contains 20 videos of shooting drills, 20 videos of armed robberies, and 10 videos of people making calls. The testing part contains 48 videos of 20 frames ($720\times720$). Videos contained in the testing dataset include phone calls, gun reviews, shooting drills, people making calls, and armed robberies at convenience stores. This dataset is labeled with the bounding boxes of hands, phones, and guns. And is available in google drive in the following link: \href{https://drive.google.com/file/d/1hF7Vr6g0fG56Oy3Jdnm2t9Y3TK9W9bn4/view?usp=sharing}{\textbf{THGP}} or go to the next url: \url{https://drive.google.com/file/d/1hF7Vr6g0fG56Oy3Jdnm2t9Y3TK9W9bn4}.

\subsection{Experiments}

To build a temporal architecture, the first step is to implement and train an image object detector with a static dataset. Then, we extend it with RNN modules, and freeze the weights from the backbone to the neck. Finally, train only the RNN modules and the head with a temporal dataset. In the following section we describe the process of training our static model (Yolov5), and our temporal architecture (TYolov5) for handgun object detection.

\subsubsection{Hardware and Software Setup}
For training our models (Yolov5 and TYolov5), we used Google Colab as GPU provider. We trained and tested our models with a Tesla V100, and Pytorch 1.7. The hardware specification of the virtual machine provided by Google Colab was an Intel(R) Xeon(R) CPU of 2.00GHz, and a RAM of 25GB.

\subsubsection{Yolov5}

To train Yolov5 with our custom dataset, we applied transfer learning from the weights trained for COCO Dataset. Then, we trained our network for 200 epochs. After each training epoch, we performed a testing routine to visualize the mean average precision (mAP) at each iteration. We save the model which obtained the highest mAP. 
For training our model we used our HGP dataset. Since we are attempting to build a static object detector, we mixed HGP with THGP dataset to increase the number of images and help the network to generalize better. Resulting in 6989 images for training, and 960 frames for testing.

\begin{figure}[!ht] 
    \centering
    \includegraphics[width=0.4\textwidth]{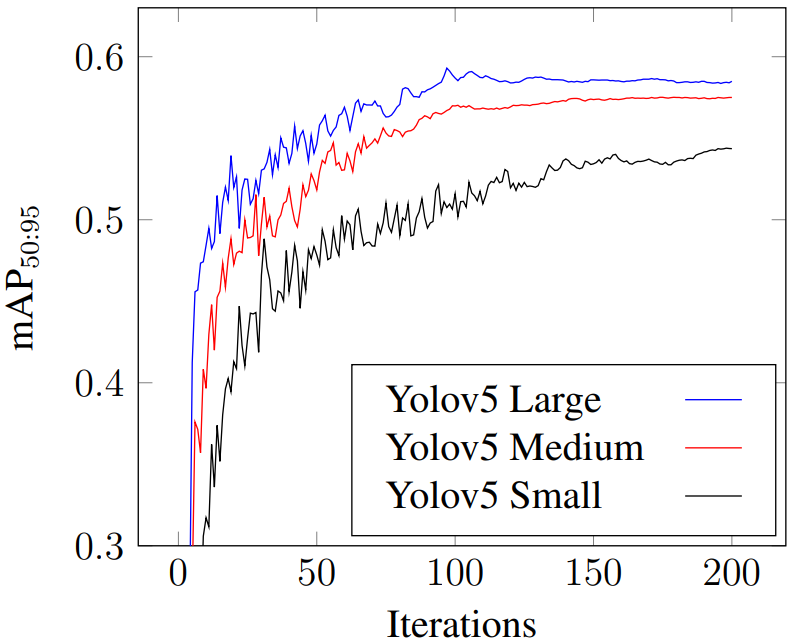}
    \caption{mAP$_{50:95}$ through iterations for each one of our models, trained 200 epochs, mixing the datasets HGP and THGP.}
    \label{fig:results_yolov5_map5090}
\end{figure}

Yolov5 stabilizes after 150 iterations (Fig. \ref{fig:results_yolov5_map5090}). Each model (Small 3.39 Hours, Medium 4.17 Hours, Large 5.98 Hours) took less than 6 hours to complete a training of 200 iterations. Results can be visualized in table \ref{table:results_TYolov5}.  


\begin{table}[!ht]
\caption{The best score obtained by Yolov5 in our HGP dataset, frames per second and the type of variables used for inference.}
    \begin{center}
         \begin{tabular}{@{}llllllll@{}}
         \toprule
          &  Methods & mAP$_{50}$ & mAP$_{50:95}$ & Var Type & FPS \\ \midrule
         A   & Yolov5s &  87.9 &   54.8 & Float 16 &   78.7 \\
         B   & Yolov5M &  90.1 &   58.2 & Float 16 &   68.49 \\
         C   & Yolov5L &  90 &   59.3  & Float 16 &  56.8 \\
         \bottomrule
        \end{tabular}
    \end{center}

\label{table:results_TYolov5}
\end{table}

The accuracy of each of our models is measured with two metrics: mAP (threshold of 50) and mAP (threshold from 50 to 95, with a step of 5). In table \ref{table:results_TYolov5}, we present the FPS, by adding the time taken by the model with a batch of size one, plus the time of the NMS algorithm.


\subsubsection{TYolov5}
To develop TYolov5, we applied transfer learning from the previously trained handgun detector with Yolov5. Next, we freezed every module from the backbone to the neck, and we trained our network for 200 epochs. After each training epoch, we performed a testing routine to visualize the mAP at each iteration. We save the model which reached the best mAP at each iteration. 

For training our temporal architectures, we combined THGP and HGP dataset. Since spatio-temporal modules quickly tend to overfit on small datasets. This resulted in a dataset which contains 6989 images for training, and 960 frames for testing. 


\subsubsection{TYolov5 with Temporal Data Augmentation}

Data augmentation might be effective alone, but does not always produce better results when combined with other techniques. To determine the best combination, we tested each temporal data augmentation technique individually. Then, repeated the same procedure by combining each data augmentation with the technique which produced the best result in the last iteration.

After running several iterations (Table \ref{table:temporal_augmentation_exploration}), we found that combining Temporal Mosaic, Blur, Temporal MixUp and Random Erasing increased the accuracy of our architecture by 2.1 points, without increasing the complexity of the model. The most notable approach was Temporal Mosaic augmentation, since by implementing it we increased the accuracy of the model by 1.4.

\begin{table}[!ht]
    \caption{Temporal Data Augmentation Techniques Exploration. The best results were obtained by combining: Temporal Mosaic, Temporal MixUp, Blur, and Random Erasing. This combination increased the accuracy of our model by 2.1.}
    \begin{center}
         \begin{tabular}{@{}lll@{}}
         \toprule
          &  Methods & mAP$_{50:95}$  \\ \midrule
         A   & Yolov5S + ConvLSTM  &  54  \\
         B & A + T. Mosaic  &   \textbf{55.4}  \\
         C & A + Blur  &  54.4  \\
         D & A + Random Erasing  &  53.5 \\
         E & A + T. MixUp  &  54.3  \\
         F & A + Gaussian Noise &  53.6  \\
         G & B + Blur  &   \textbf{55.6}  \\
         H & B + Random Erasing  &  54.8  \\
         I & B + T. MixUp  & 55.1  \\
         J & B + Gaussian Noise  & 54.3  \\
         K & G + Random Erasing  &   56  \\
         L & G + T. MixUp &  \textbf{56.1}  \\
         M & G + Gaussian Noise & 55.3  \\
         N & L + Random Erasing &   \textbf{56.1}  \\
         P & L + Gaussian Noise & 55.1  \\
        \bottomrule
        \end{tabular}
    \end{center}
\label{table:temporal_augmentation_exploration}
\end{table}

\subsubsection{TYolov5 Small}

\begin{figure}[!ht] 
    \centering
    \includegraphics[width=0.4\textwidth]{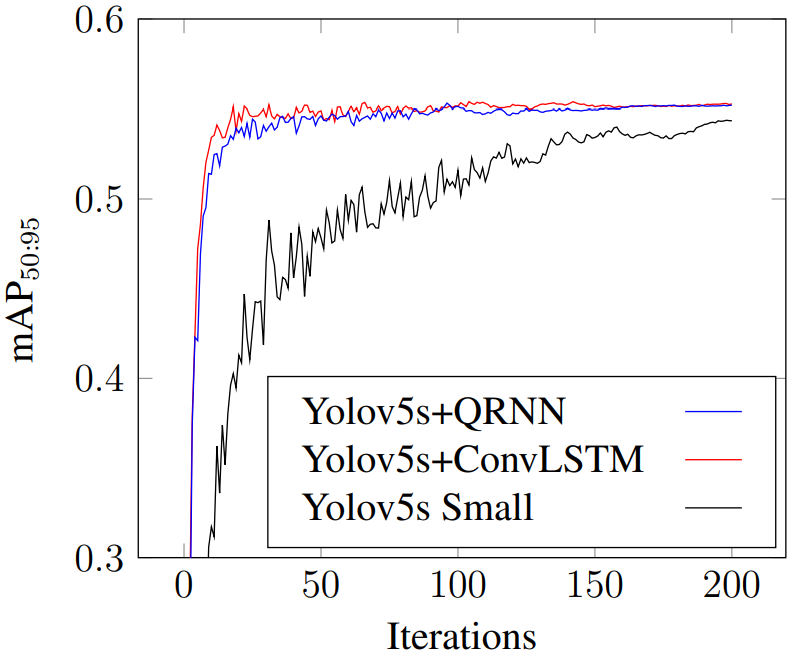}
    \caption{mAP$_{50:95}$ accuracy through iterations for each one of our models, trained 200 epochs, mixing the datasets HGP and THGP.}
    \label{fig:results_Tyolov5s_map5095}
\end{figure}

To train the small TYolov5, we used the best data augmentation combination defined in table \ref{table:temporal_augmentation_exploration}. Then we trained the network in THGP+HGP for 200 iterations. The mAP$_{50:95}$ score obtained at each iteration is visualized in Fig. \ref{fig:results_Tyolov5s_map5095}.
Results for our smaller model can be visualized in table \ref{table:TYoloS}.

For the small architecture, both architectures offered comparable results.  QRNNs got better mAP than ConvLSTMs, but the difference in  mAP$_{50:95}$ was just 0.3. However, QRNNs presented a faster inference time than ConvLSTMs.

\begin{table}[!ht]
    \caption{TYolo Small. The accuracy of TYolov5s+QRNNs improved the results obtained by Yolov5s.}
    \begin{center}
         \begin{tabular}{@{}llllll@{}}
         \toprule
          &  Methods & mAP$_{50}$ & mAP$_{50:95}$  & Var Type & FPS \\ \midrule
         A& Yolov5s &  87.9 &   54.8 &  Float 16 &   \textbf{78.7} \\
         B& A+ConvLSTM  &   \textbf{88.5} &   55.6  &  Float 32 &  50 \\
         C& A+QRNN &  88.3 &  \textbf{55.9} & Float 32 &  52.6 \\
          \bottomrule

        \end{tabular}
    \end{center}

\label{table:TYoloS}
\end{table}

\subsubsection{TYolov5 Medium}

\begin{figure}[!ht] 
    \centering
    \includegraphics[width=0.4\textwidth]{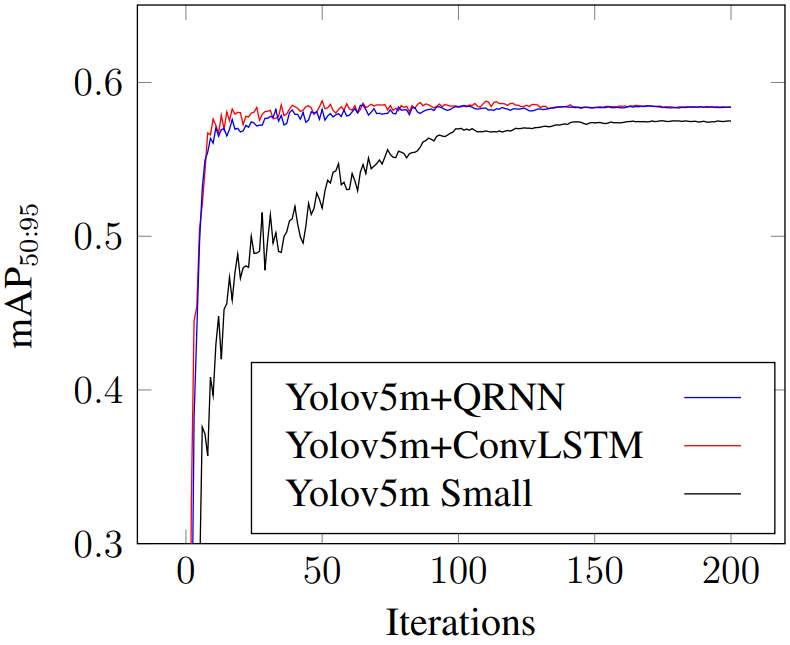}
    \caption{mAP$_{50:95}$ accuracy through iterations for each one of our models, trained 200 epochs, mixing the datasets HGP and THGP.}
    \label{fig:results_Tyolov5m_map5095}
\end{figure}

For training medium TYolov5, we used the best data augmentation combination defined in table \ref{table:temporal_augmentation_exploration}. Then we trained the network in THGP for 200 iterations. The mAP$_{50:95}$ score obtained at each iteration is visualized in Fig. \ref{fig:results_Tyolov5m_map5095}.
Results for our medium model can be visualized in table \ref{table:TYoloM}.

For the medium architecture, ConvLSTMs offered better results than QRNNs. Nevertheless, the difference in mAP$_{50:95}$ was just 0.7. However, QRNNs still presented a faster inference time than ConvLSTMs.

\begin{table}[!ht]
\caption{TYolo Medium. The accuracy of TYolov5m+QRNNs improved the results obtained by Yolov5m.}
    \begin{center}
         \begin{tabular}{@{}llllll@{}}
         \toprule
          &  Methods & mAP$_{50}$ & mAP$_{50:95}$  & Var Type & FPS \\ \midrule
         A   &Yolov5M &  90.1 &   58.2  & Float 16 &  \textbf{68.49} \\
         B & A+ConvLSTM  &   \textbf{90.7} &   \textbf{59.7} & Float 32 &   29.5 \\
         C & A+QRNN &  \textbf{90.7} &  59 & Float 32 & 31.0 \\
          \bottomrule
        \end{tabular}
    \end{center}
\label{table:TYoloM}
\end{table}

\subsubsection{TYolov5 Large}

\begin{figure}[!ht] 
    \centering
    \includegraphics[width=0.4\textwidth]{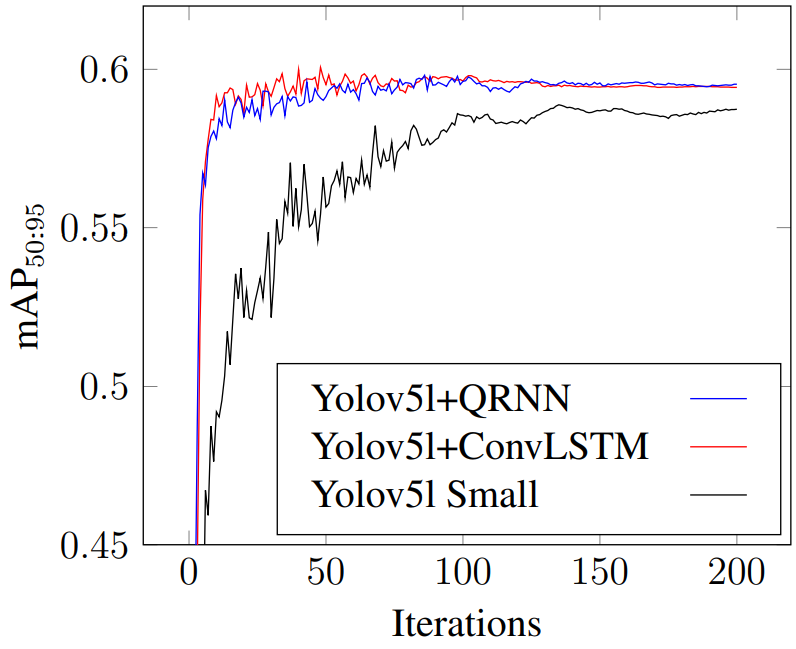}
    \caption{mAP$_{50:95}$ accuracy through iterations for each one of our models, trained 200 epochs, mixing the datasets HGP and THGP.}
    \label{fig:results_TYolov5l_map5095}
\end{figure}

For training large TYolov5, we used again the best data augmentation combination defined in table \ref{table:temporal_augmentation_exploration}. We trained the network in THGP for 200 iterations. The mAP$_{50:95}$ score obtained at each iteration is visualized in Fig. \ref{fig:results_TYolov5l_map5095}.
Results for our large model can be visualized in table \ref{table:TYoloL}. 

For the large TYolov5 architecture, QRNNs offered better results.  Nevertheless, the difference in mAP$_{50:95}$ between both architectures was just 0.2. QRNNs still presented a faster inference time than ConvLSTMs.

\begin{table}[!ht]
    \caption{TYolo Large. The accuracy of TYolov5l+QRNNs improved the results obtained by Yolov5l.}
    \begin{center}
         \begin{tabular}{@{}llllll@{}}
         \toprule
          &  Methods & mAP$_{50}$ & mAP$_{50:95}$  & Var Type & FPS \\ \midrule
         A   & Yolov5L &  90 &   59.3 & Float 16 &  \textbf{56.8} \\
         B & A+ConvLSTM  &   \textbf{90.7} &   60 & Float 32 &   19.2 \\
         C & A+QRNN &  90.6 &  \textbf{60.2} & Float 32 &  22.5 \\
          \bottomrule
        \end{tabular}
    \end{center}

\label{table:TYoloL}
\end{table}

\subsection{Discussion}
QRNNs applied to the domain of handgun object detection, offer comparable results to ConvLSTMs. Nonetheless, they reduce inference time when using full precision floats. We proved that QRNNs are suitable for real-world applications, like handgun real-time detection.

Temporal data augmentation techniques increased the accuracy of the model by 2.1. This shows the importance of designing and implementing temporal data augmentation approaches for training spatio-temporal models like TYolov5. 

\section{Conclusion and future work}
\label{sc:conclusion} 

\subsection{Conclusions}

This work introduces three temporal Yolov5 architectures for temporal handgun detection based on QRNNs. The small and medium architectures operate above 30 frames per second, enough to be categorized as real-time detectors. Each architecture offers comparable results to networks based on ConvLSTMs. Nonetheless, QRNNs reduced the inference time when compared to ConvLSTMs, presenting a superior FPS for all architectures.

Additionally, we introduced two publicly available datasets labeled with the bounding boxes of hands, guns, and phones. One has 2199 static images, and another with 5960 frames of videos. Most of the datasets used in state-of-the-art for handgun detection, contain images of guns without context. Our datasets contain images and videos of objects used in real-world scenarios, which helps the models to generalize better.

Temporal data augmentation improved the accuracy of TYolov5 in our temporal dataset, with the benefit of not increasing the complexity of the model, or the number of parameters. Temporal data augmentation has not received the same attention as static data augmentation techniques, which makes temporal mosaic and temporal mixup worth exploring.

Due to its restrictions, designing real-time causal object detectors is a challenging task in computer vision. We hope our proposed datasets, architectures, and data augmentation techniques, serve as a basis towards building more robust and faster approaches.
 
\subsection{Future Work}
QRNNs showed comparable results to Convolutional LSTMs in our temporal dataset. However, proving that this architecture can be generalized to more object classes, is worth researching. Moreover, it is worth testing which are the generalization capabilities of QRNNs to other tasks like action recognition.

\section*{acknowledgment}
To Glenn Jocher, for the help during the development of TYolov5 and the idea of Mosaic data augmentation.

Funding: This work was supported by Tecnologico de Monterrey and Consejo Nacional de Ciencia y Tecnologia (CONACYT)

\bibliographystyle{unsrt}  
\bibliography{refs}

\end{document}